\documentclass{article}

\PassOptionsToPackage{numbers}{natbib}

\usepackage[preprint]{neurips_2024}

\usepackage[utf8]{inputenc} 
\usepackage[T1]{fontenc}    
\usepackage[colorlinks=true, citecolor=violet]{hyperref}
\usepackage{url}            
\usepackage{booktabs}       
\usepackage{amsfonts}       
\usepackage{nicefrac}       
\usepackage{microtype}      
\usepackage{xcolor}         
\usepackage{graphicx}
\usepackage{multirow}
\usepackage{ltablex}
\keepXColumns
\usepackage{tabularx}
\usepackage{caption}
\usepackage{etoolbox}
\usepackage{subcaption} 

\usepackage{longtable}
\usepackage{booktabs}
\usepackage{pdflscape}
\usepackage{tcolorbox}

\makeatletter

\title{Prompting language influences diagnostic reasoning and accuracy of large language models}

\author{%
  Adrien Bazoge \quad Josselin Corvellec \\ 
  \textbf{Sofiane Djillali Sid-Ahmed} \quad \textbf{Pierre-Antoine Gourraud} \\
  Data Clinic, Nantes University Hospital, France\\
}

\begin{document}

\maketitle

\begin{abstract}
Large language models (LLMs) are increasingly explored for clinical decision support, yet most evaluations are conducted in English, leaving their reliability in other languages uncertain. Here we evaluate the impact of prompting language on diagnostic reasoning and final diagnosis accuracy by comparing English and French performance across five LLMs (o3, DeepSeek-R1, GPT-4-Turbo, Llama-3.1-405B-Instruct, and BioMistral-7B). A total of 180 clinical vignettes covering 16 medical specialties were assessed by two physicians using an 18-point scale evaluating both diagnosis accuracy and reasoning quality. Four of the five models performed better in English (mean difference 0.37–0.91, adjusted \textit{p}~$<$~0.05), with the gap spanning multiple aspects of reasoning, including differential diagnosis, logical structure, and internal validity. o3 was the only model showing no overall language effect. These findings demonstrate that prompting language remains a critical determinant of LLM clinical performance, with implications for equitable linguistico-cultural deployment worldwide.

\end{abstract}

\section{Introduction}

Among AI applications, the accessibility of Large Language Models (LLMs) is rapidly transforming many sectors including medicine, with promising applications in assisting clinicians in diagnostic reasoning and decision-making, reducing the burden of administrative tasks to free up medical time, and enabling people to easily access basic health advice~\cite{Topol2019,doi:10.1056/AIcs2400420,Menezes2025}. By encoding vast clinical knowledge and analyzing complex information from electronic health records at scale and in near real time, LLMs have the potential to support clinicians across specialties and healthcare systems, while also improving access to health information and high-quality care, especially in low-resource settings~\cite{arora2025healthbenchevaluatinglargelanguage,ahsan2024retrieving,Singhal2023,nori2023capabilitiesgpt4medicalchallenge}.

Despite this potential, important challenges remain before LLMs can be responsibly integrated into clinical practice, ensuring that innovation aligns with Hippocratic values—prioritizing patient safety, autonomy, privacy, and equal opportunity for quality care~\cite{Clusmann2023}. Most models are trained predominantly on English-language data from high-income regions like the United States and Europe, and primarily benchmarked in English-speaking contexts, raising concerns about their reliability in multilingual and culturally diverse healthcare settings~\cite{durmus2024towards,myung2024blend}. This is particularly acute in medicine, where language is inseparable from culture, epidemiology, clinical practice, and regulation. Substantial amounts of clinically relevant knowledge remain inaccessible, underrepresented, or unavailable in training corpora due to privacy constraints. Even when such knowledge is encoded, its effective use depends on its accurate interpretation and application within the user's own linguistic, cultural, and healthcare context. Without robust multilingual and multicultural alignment at both the training and deployment stages, LLMs risk missing the opportunity to reduce inequities in access to trustworthy health information, reliable clinical decision support, and contextually appropriate care, especially for populations whose languages and healthcare realities are poorly represented in the data that shape these models. Furthermore, evaluations of LLMs mostly rely on simplified standardized and multiple-choice benchmarks~\cite{Singhal2023,Singhal2025}, which do not capture the complexity of real clinical routine practice. As a result, their evaluation in real-world clinical settings remains limited~\cite{Menezes2025,arora2025healthbenchevaluatinglargelanguage}, leaving uncertainties about their reliability and safety across diverse healthcare environments.

Prior work has demonstrated the ability of LLMs to generate accurate diagnoses and differential diagnoses from simulated clinical vignettes and sequential clinical encounters, but most evaluations have been conducted exclusively in English~\cite{10.1001/jama.2023.8288,Savage2024,10.1001/jamainternmed.2023.2909,nori2025sequentialdiagnosislanguagemodels,Sandmann2025}. The few studies that have explored multilingual evaluation in medical context were either limited in the number of cases~\cite{Menezes2025}, restricted to multiple-choice format from medical examinations~\cite{ALONSO2024102938,qiu2024towards,strasser2026performance,yang2026toward}, or did not explicitly analyze performance differences across languages~\cite{arora2025healthbenchevaluatinglargelanguage}. Emerging evidence from these studies indicates that model performances can vary significantly across languages~\cite{Menezes2025,ALONSO2024102938,qiu2024towards,strasser2026performance,yang2026toward,10.1145/3589334.3645643}, raising the question of whether performance on clinical decision tasks, including diagnostic reasoning, remains consistent in non-English contexts.

To address this gap, we designed a bilingual English-French comparative evaluation framework of diagnostic reasoning and final diagnosis accuracy, as these represent the entry door of clinical decision-making and are crucial for delivering effective patient care~\cite{10.1001/jamahealthforum.2021.2430,10.1001/jama.2021.22396}. Five LLMs spanning different architectures and capability levels were evaluated: o3~\cite{jaech2024openai}, DeepSeek-R1~\cite{Guo2025}, GPT-4-Turbo~\cite{openai2024gpt4technicalreport}, Llama-3.1-405B-Instruct (Llama-405B)~\cite{grattafiori2024llama3herdmodels}, and BioMistral-7B~\cite{labrak-etal-2024-biomistral}. A total of 180 clinical vignettes covering 16 medical specialties and multiple diagnostic and reasoning types were independently evaluated by two physicians using an 18-point scale assessing both the accuracy of the final diagnosis and the quality of the underlying clinical reasoning.

\section{Methods}

\subsection{Design of the vignettes}

This study investigates diagnostic reasoning and clinical diagnosis across medical and medical-surgical specialties. Disciplines with a primarily technical focus (radiology, clinical biology, anatomical pathology and nuclear medicine) and purely surgical fields were excluded. Sixteen specialties were retained, covering the breadth of clinical medicine: emergency and critical care, endocrinology and metabolism, gynecology, oncology-hematology, hepatology-gastroenterology, infectious and tropical diseases, cardiovascular medicine, general practice, internal medicine, neurology, head and neck medicine, pediatrics, pulmonology, psychiatry, rheumatology, and urology-nephrology.
For each specialty, approximately 10 vignettes were developed (range: 6–17), except for general practice, which was the specialty of the two physicians, for which 32 vignettes were created, yielding a total of 180 vignettes. Each vignette was designed with a predefined expected diagnosis and contained all information necessary to establish it, including first-line ancillary test results when clinically appropriate. We adopted this vignette-based design to approximate the realities of early clinical encounters, such as those in outpatient or emergency care. In these settings, physicians usually work with a narrow set of focused questions, resulting in histories that may be sparse, incomplete, or occasionally contain irrelevant details.

Vignettes were derived from three sources: (1) synthetic vignettes created de novo by physicians in accordance with current medical guidelines, (2) vignettes adapted from reference materials (textbooks, lectures, and residency training resources), and (3) non-identifying details based on real-world clinical encounters. Vignettes adapted from reference texts were reformulated into vignette format to meet predefined criteria and to minimize potential overlap with models' training data available online. For synthetic and real vignettes, final diagnoses were proposed by physicians; for vignettes based on reference texts, diagnoses were extracted from the source. Diagnoses were considered the expected diagnosis, regardless of certainty, and reflected the most relevant, probable, or clinically actionable condition. For example, in a critically ill patient, the expected diagnosis was septic shock rather than identification of the specific pathogen. Depending on context, diagnoses could be etiological (\textit{n}~=~113), syndromic (\textit{n}~=~54), or paraclinical (\textit{n}~=~13).

For all vignettes, physicians provided a reference diagnostic reasoning pathway leading to the final diagnosis. Reasoning processes were categorized into five types, each being assigned a single predominant type for analytical purposes, although clinical reasoning in practice often involves multiple simultaneous approaches:
\begin{itemize}
    \item Case recognition (\textit{n}~=~57): non-analytical reasoning based on pattern recognition or recall of previously encountered cases; effective for simple and typical cases, and requiring a solid clinical background.
    \item Hypothetico-deductive (\textit{n}~=~37): systematic evaluation of diagnostic hypotheses, often generated through intuitive pattern recognition, and tested via history-taking, clinical examination, and ancillary tests to confirm or exclude potential diagnoses.
    \item Forward chaining (\textit{n}~=~55): an analytical process moving stepwise from clinical and ancillary results to diagnosis by applying causal or conditional rules (clinical knowledge, pathophysiology, etc.).
    \item Algorithmic (\textit{n}~=~14): a binary, stepwise process in which the physician arrives at a diagnosis by successive exclusions, depending on the presence or absence of signs or the positivity or negativity of tests.
    \item Probabilistic (\textit{n}~=~17): estimation of post-test diagnostic probability using prevalence data in the patient's population and likelihood ratios for clinical and ancillary results; particularly suited for contexts of diagnostic uncertainty.
\end{itemize}

All 180 vignettes were constructed in French. Each vignette, along with the corresponding diagnostic reasoning and final diagnosis, was independently translated into English by both physicians. Cross-verification of translations was performed across all vignettes to ensure clinical accuracy and semantic equivalence.

\subsection{Models selection and prompting}

Five large language models were evaluated, selected to represent a range of architectures, training approaches, and accessibility levels. The study was conducted in two phases. In the first phase (August-September 2024), three models were evaluated: GPT-4-Turbo via the OpenAI API on 6 August 2024, Llama-3.1-405B-Instruct via the NVIDIA API on 10 September 2024, and BioMistral-7B, deployed locally on NVIDIA A100 80GB GPU. In a second phase (July 2025), two additional, more recent models with improved reasoning capabilities were included: DeepSeek-R1 via the DeepSeek API and OpenAI o3 via the OpenAI API, both queried on 29 July 2025.

Each model was queried once per vignette in each language, with default generation parameters. The prompt was identical across models, with only the language instruction and vignette language varying between English and French:

\begin{tcolorbox}[colback=gray!10, colframe=gray!70, title=Prompt template, fonttitle=\bfseries\small, fontupper=\small\ttfamily]
Act as a doctor taking a medical history from a patient. Try your best to give your clinical reasoning in order to make an accurate final diagnosis. Your response must be written in \{language\}. Format the output as json with two fields: `diagnostic' with your final diagnosis and `clinical\_reasoning' with your clinical reasoning that led to this diagnosis. \{vignette\}
\end{tcolorbox}

\noindent where \texttt{\{language\}} was replaced by ``English'' or ``French'' and \texttt{\{vignette\}} by the corresponding vignette text in English or in French.

\subsection{Evaluation framework}

Model outputs were assessed using an 18-point evaluation scale based on six criteria (detailed scoring scales are provided in Supplementary Table~\ref{tab:table9}). Five criteria evaluated the quality of the diagnostic reasoning:
\paragraph{Internal validity (0 to 5 score)} This criterion assesses the ability to extract and interpret relevant data from the vignette. A high-quality response requires inclusion of all elements from the vignette that are useful for diagnosis, as well as the interpretation of ancillary findings that contribute to reasoning. This item also evaluates whether descriptive findings were translated into appropriate medical terminology (e.g., "purplish skin lesion not blanching under pressure" should prompt the recognition of purpura). Similarly, numerical clinical data were expected to be contextualized, such as defining 38.5°C as fever, systolic blood pressure above a threshold as hypertension, or body mass index $\geq$ 25 kg/m$^2$ as overweight. The same principle was applied to ancillary tests: when not pre-interpreted in the vignette, we expected them to be integrated whenever relevant for diagnosis.
\paragraph{External validity (0 to 3 score)} This criterion evaluates the scientific accuracy of medical knowledge introduced by the model beyond the vignette description. If no external knowledge was invoked, a default score of 3 was assigned to not penalize the absence of such content that is often unnecessary to the diagnosis.
\paragraph{Hypotheses and differential diagnosis (0 to 1 score)} This criterion assesses the ability to generate relevant hypotheses and differential diagnoses, ideally organized by likelihood or severity. Even if a differential diagnosis could be easily excluded, its mention often enriches clinical reasoning. If differentials were cited in the physicians’ diagnostic reasoning but omitted by the model, a score of 0 was assigned. Conversely, for vignettes in which no differential diagnoses were expected, full points were awarded to avoid penalizing otherwise valid reasoning.
\paragraph{Logical structure (0 to 4 score)} This criterion measures the coherence and organization of the reasoning, independently of content accuracy. Particular attention was paid to logical order of presentation (e.g., clinical data usually preceding ancillary results), the absence of contradictions, and the avoidance of irrelevant assertions.
\paragraph{Expression (0 to 2 score)} Responses were penalized for errors in expression, meaning or syntax. Answers generated in English when French was expected were also downgraded.

The sixth criterion, accuracy of final diagnosis (0 to 3 score), compared the model’s diagnosis with the reference diagnosis established by physicians.
Two physicians, blinded to each other’s assessments, independently assessed all model outputs for both languages across all 180 vignettes, yielding 360 paired assessments per model.

\subsection{Statistical analysis}

The primary analysis compared English and French performance within each model using linear mixed models (LMM) with language as a fixed effect and vignette and two raters as random intercepts. All scores, including ordinal and binary sub-scores, were treated as continuous in the LMM. One-sided tests were used to evaluate the hypothesis that models perform better in English than in French. p-values were adjusted for multiple comparisons using Bonferroni correction ($k~=~5$). Results are reported as mean difference (EN – FR) with 95\% confidence intervals (Wald method). Descriptive statistics are reported as median [interquartile range] and percentage of observations achieving the maximum score. 

Inter-rater reliability was assessed using the intraclass correlation coefficient (ICC, two-way random, single measures, absolute agreement) for the overall score and quadratic-weighted Cohen’s kappa for ordinal sub-scores, with unweighted Cohen’s kappa for the binary differential diagnosis criterion. 95\% confidence intervals were computed by bootstrap (2,000 resamples).

Residual normality was assessed using the Shapiro-Wilk test; departures from normality were observed for all models (all \textit{p}~<~0.001). As a sensitivity analysis, scores from both raters were averaged per vignette (\textit{n}~=~180), and paired Wilcoxon signed-rank tests were performed on the aggregated scores with the same one-sided alternative and Bonferroni correction. Spearman rank correlations were computed between diagnostic reasoning scores and final diagnosis accuracy for each model and language. Differences between English and French correlations were tested using Fisher’s z-transformation.

Overall sample size is \textit{n}~=~21,600 items evaluated (180 vignettes, 2 physicians, 5 LLMs, 2 languages, 6 score components).  All analyses were performed using R (version 4.5.1) with the lme4, lmerTest, and irr packages. Inter-rater agreement was computed using Python (version 3.11.11) with scikit-learn and SciPy.

\section{Results}

\subsection{Overall model performance}

All five models achieved median overall scores above 8/18 in both languages, though performance varied substantially across models (Table~\ref{tab:table1}, Supplementary Figure~\ref{fig:4}). o3 achieved the highest performance with a median of 18.00 [16.00, 18.00] in both English and French, followed by GPT-4-Turbo (EN: 17.00 [15.00, 18.00]; FR: 16.00 [14.00, 17.00]) and Llama-405B (EN: 17.00 [15.00, 18.00]; FR: 16.00 [13.00, 17.00]). DeepSeek-R1 performed similarly (EN: 16.00 [15.00, 18.00]; FR: 16.00 [15.00, 17.00]), while BioMistral-7B scored substantially lower (EN: 9.00 [6.75, 11.00]; FR: 8.00 [6.00, 10.00]), despite being specialized in the medical domain, highlighting the greater difficulty that smaller models face with complex diagnostic tasks.

\begin{table}[htbp]
\centering

\resizebox{\textwidth}{!}{%
\begin{tabular}{lccccccc}
\toprule
\textbf{Model} & \textbf{ICC EN [95\% CI]} & \textbf{EN median [IQR]} & \textbf{ICC FR [95\% CI]} & \textbf{FR median [IQR]} & \textbf{Mean diff} & \textbf{95\% CI} & \textbf{Adj. \textit{p}} \\
 & & & & & \textbf{(LMM)} & & \\
\midrule
o3 & 0.22 [0.09, 0.35] & 18.00 [16.00, 18.00] & 0.15 [0.04, 0.26] & 18.00 [16.00, 18.00] & 0.08 & [-0.12, 0.27] & 1 \\
\addlinespace[3pt]
DeepSeek-R1 & 0.37 [0.22, 0.50] & 16.00 [15.00, 18.00] & 0.40 [0.22, 0.54] & 16.00 [15.00, 17.00] & 0.37 & [0.10, 0.64] & 0.0207 \\
\addlinespace[3pt]
GPT-4-Turbo & 0.31 [0.17, 0.45] & 17.00 [15.00, 18.00] & 0.52 [0.39, 0.63] & 16.00 [14.00, 17.00] & 0.49 & [0.25, 0.73] & $<$0.001 \\
\addlinespace[3pt]
Llama-405B & 0.61 [0.49, 0.70] & 17.00 [15.00, 18.00] & 0.56 [0.44, 0.65] & 16.00 [13.00, 17.00] & 0.91 & [0.66, 1.17] & $<$0.001 \\
\addlinespace[3pt]
BioMistral-7B & 0.82 [0.76, 0.86] & 9.00 [6.75, 11.00] & 0.84 [0.78, 0.88] & 8.00 [6.00, 10.00] & 0.78 & [0.28, 1.28] & 0.00611 \\
\bottomrule
\end{tabular}
}
\vspace{4pt}
\caption{\textbf{Performance comparison between English and French on the overall score (0--18) for each model.} Inter-rater reliability is reported as the intraclass correlation coefficient (ICC, two-way random, single measures, absolute agreement) with bootstrap 95\% confidence intervals. Scores are reported as median [interquartile range]. Differences were assessed using linear mixed models with language as fixed effect, vignette and rater as random intercepts. P-values are one-sided (EN > FR) and Bonferroni-adjusted ($k = 5$).}
\label{tab:table1}
\end{table}

\subsection{Inter-rater agreement}

Inter-rater agreement statistics between the two physicians across models and languages are presented in Table~\ref{tab:table1}. The overall inter-rater agreement was fair (mean ICC~=~0.48). Agreement was highest for BioMistral-7B (ICC~=~0.82–0.84, score range: median 8.00–9.00, IQR 6.00–11.0) and lowest for o3 (ICC~=~0.15–0.22, median 18.00, IQR 16.00–18.00). GPT-4-Turbo (ICC~=~0.31–0.52), DeepSeek-R1 (ICC~=~0.37–0.40) and Llama-405B (ICC~=~0.56–0.61) showed intermediate levels. Agreement was generally higher in French than in English across models. Among individual criteria, agreement was highest for final diagnosis accuracy (mean weighted $\kappa$~=~0.61) and lowest for expression (mean weighted $\kappa$~=~0.15) and differential diagnosis (mean $\kappa$~=~0.12). Detailed inter-rater agreement results are provided in Supplementary Table~\ref{tab:table6}.

\begin{figure}[htbp]
\centering
\includegraphics[width=\linewidth]{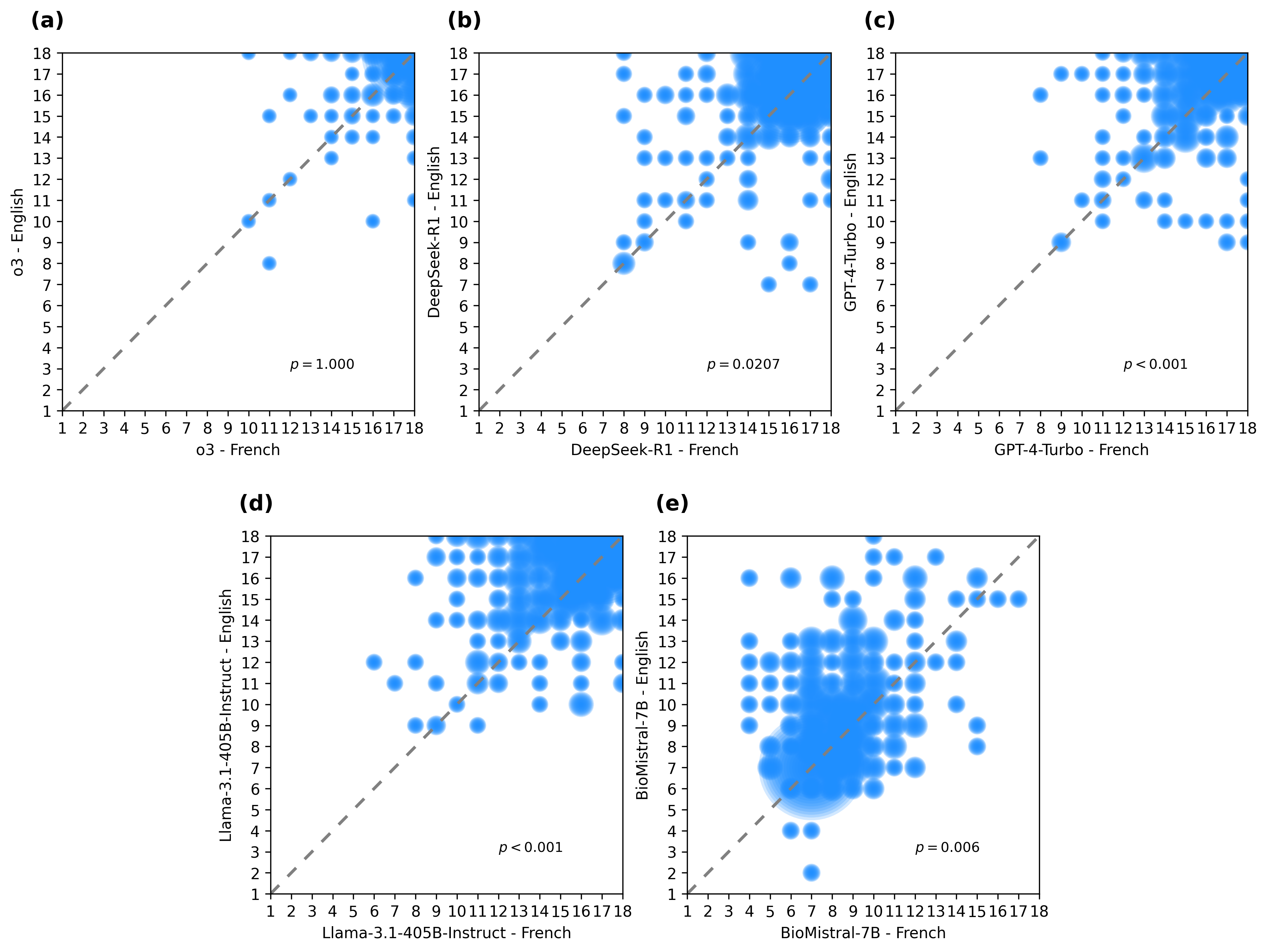}
\vspace{4pt}
\caption{\textbf{Pairwise comparison of model performance between English and French prompting.} Bubble plots showing the results of 360 pairwise comparisons on an 18-point scale, with two physicians each independently assessing all 180 vignettes, comparing English versus French prompting: \textbf{(a)} o3 (linear mixed model, language as fixed effect, vignette and rater as random intercepts, one-sided test EN > FR, Bonferroni correction $k~=~5$, mean difference~=~0.08, 95\% CI [$-$0.12, 0.27], adjusted \textit{p}~=~1.000); \textbf{(b)} DeepSeek-R1 (mean difference~=~0.37, 95\% CI [0.10, 0.64], adjusted \textit{p}~=~0.021); \textbf{(c)} GPT-4-Turbo (mean difference~=~0.49, 95\% CI [0.25, 0.73], adjusted \textit{p}~<~0.001); \textbf{(d)} Llama-405B (mean difference~=~0.91, 95\% CI [0.66, 1.17], adjusted \textit{p}~<~0.001); \textbf{(e)} BioMistral-7B (mean difference~=~0.78, 95\% CI [0.28, 1.28], adjusted \textit{p}~=~0.006).}
\label{fig:1}
\end{figure}

\subsection{English vs French comparison}

\begin{figure}[htbp]
\centering
\includegraphics[width=\linewidth]{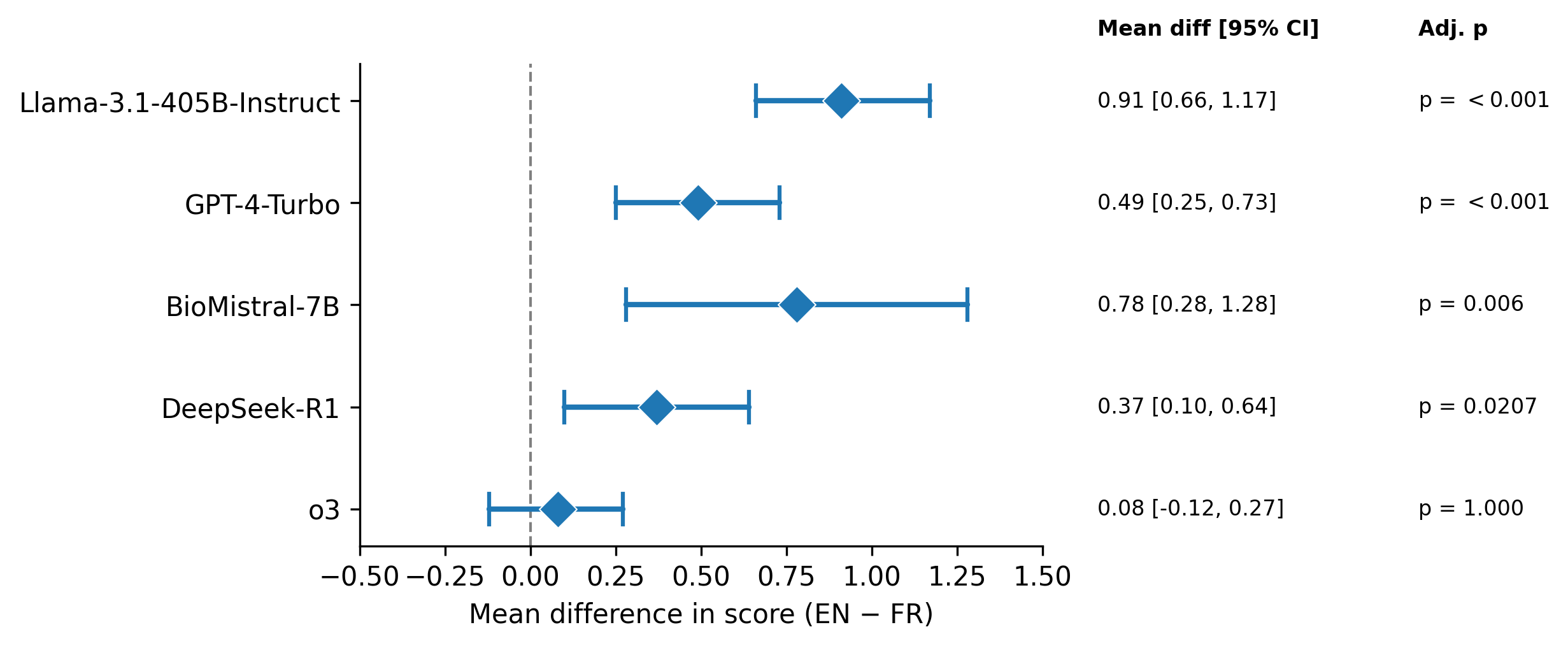}
\caption{\textbf{Effect of prompting language on overall model performance.} Forest plot of mean differences (EN – FR) in overall score (0–18) with 95\% confidence intervals from linear mixed models. Bonferroni-adjusted P-values are shown (k~=~5, one-sided test EN~>~FR).}
\label{fig:2}
\end{figure}

Four of the five models performed significantly better in English than in French (Table~\ref{tab:table1}, Figure~\ref{fig:1}, Figure~\ref{fig:2}). The largest language gap was observed for Llama-405B (mean difference~=~0.91, 95\% CI [0.66, 1.17], adjusted \textit{p}~<~0.001) followed by BioMistral-7B (0.78 [0.28, 1.28], \textit{p}~=~0.006), GPT-4-Turbo (0.49 [0.25, 0.73], p~<~0.001), and DeepSeek-R1 (0.37 [0.10, 0.64], \textit{p}~=~0.021). o3 was the only model showing no performance difference between languages (0.08 [-0.12, 0.27], \textit{p}~=~1.000). Sensitivity analyses using paired Wilcoxon signed-rank tests on aggregated scores (\textit{n}~=~180) confirmed these findings (Supplementary Table~\ref{tab:table7}).

\subsection{Language comparison across evaluation criteria}

\begin{figure}[htbp]
\centering

\includegraphics[width=\linewidth]{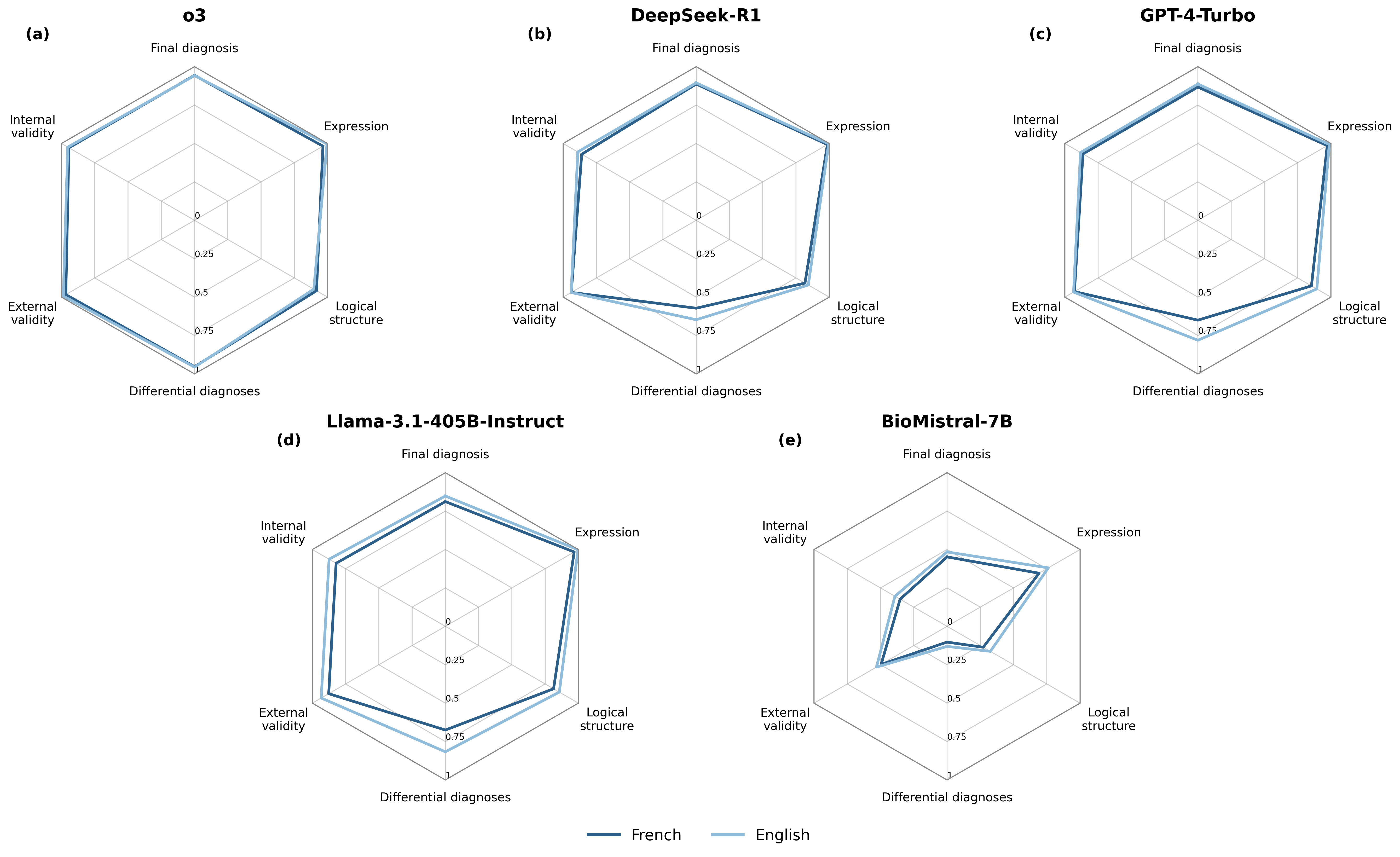}
\caption{\textbf{Detailed performance across evaluation criteria by language.} Radar plots of mean performance across six evaluation criteria for each model in French (dark blue) and English (light blue). All scores are normalized to a 0--1 scale representing the proportion of the maximum score achieved (original scales: final diagnosis 0--3, internal validity 0--5, external validity 0--3, differential diagnosis 0--1, logical structure 0--4, expression 0--2).}
\label{fig:3}
\end{figure}

Analysis of individual evaluation criteria showed that the language gap was not uniform across dimensions of clinical reasoning (Table~\ref{tab:table2}, Figure~\ref{fig:3}, Supplementary Table~\ref{tab:table8}). For differential diagnosis, performance differences favoring English were observed for Llama-405B (mean difference~=~0.14, \textit{p}~<~0.001), GPT-4-Turbo (0.13, \textit{p}~<~0.001), and DeepSeek-R1 (0.08, \textit{p}~=~0.021), with English-prompted models generating relevant differential diagnoses more consistently. o3 showed no language effect on this criterion, with near-ceiling performance in both languages (95.0\% vs 95.3\% at maximum score). For logical structure, language effects were observed for BioMistral-7B (0.21, \textit{p}~=~0.001), Llama-405B (0.17, \textit{p}~=~0.005), and GPT-4-Turbo (0.16, \textit{p}~=~0.005), indicating more coherent reasoning organization in English. For internal validity, differences were found for Llama-405B (0.26, \textit{p}~<~0.001) and DeepSeek-R1 (0.15, \textit{p}~=~0.03), suggesting better extraction and interpretation of clinical data in English. External validity showed a language gap for Llama-405B (0.17, \textit{p}~<~0.001). A small difference was also observed for o3 (0.06, \textit{p}~=~0.029), the only criterion for which o3 showed a language effect. Expression quality differed most for BioMistral-7B (0.14, \textit{p}~=~0.022), reflecting more frequent syntactic or language-compliance errors in French outputs. Small differences were also observed for o3 and Llama-405B (both 0.06, \textit{p}~<~0.001). Final diagnosis accuracy showed a language effect only for Llama-405B (0.11, \textit{p}~=~0.014), despite this model achieving a median of 3/3 in both languages.

Across models, diagnostic reasoning scores were positively correlated with final diagnosis accuracy in both languages ($\rho$ ranging from 0.47 to 0.75, all \textit{p}~<~0.001). For DeepSeek-R1, Llama-405B and GPT-4-Turbo, correlations were slightly higher in English than in French, but the differences were not significant (Fisher's z, all \textit{p}~>~0.2). In contrast, BioMistral-7B showed a significantly stronger correlation in English ($\rho$~=~0.75) than in French ($\rho$~=~0.60; Fisher's z~=~3.91, \textit{p}~<~0.001), suggesting that for this model, the alignment between diagnostic reasoning and diagnosis accuracy is more robust in English.

\begin{table}[htbp]
\centering
\resizebox{\textwidth}{!}{%
\begin{tabular}{llccccccc}
\toprule
\textbf{Criterion (scale)} & \textbf{Model} & \textbf{EN median} & \textbf{\% max} & \textbf{FR median} & \textbf{\% max} & \textbf{Mean diff} & \textbf{95\% CI} & \textbf{Adj. \textit{p}} \\
 &  & \textbf{[IQR]} & \textbf{EN} & \textbf{[IQR]} & \textbf{FR} & \textbf{(LMM)} & \ &  \\
\midrule
Final diagnosis (0--3) & o3 & 3.00 [3.00;3.00] & 85.8\% & 3.00 [3.00;3.00] & 86.4\% & -0.01 & [-0.06, 0.05] & 1 \\
 & DeepSeek-R1 & 3.00 [3.00;3.00] & 78.1\% & 3.00 [3.00;3.00] & 76.1\% & 0.02 & [-0.05, 0.09] & 1 \\
 & GPT-4 & 3.00 [3.00;3.00] & 77.8\% & 3.00 [2.00;3.00] & 71.4\% & 0.06 & [-0.02, 0.13] & 0.302 \\
 & Llama-405B & 3.00 [2.00;3.00] & 68.1\% & 3.00 [2.00;3.00] & 63.3\% & 0.11 & [0.03, 0.18] & 0.0141 \\
 & BioMistral & 1.00 [1.00;3.00] & 26.4\% & 1.00 [1.00;2.00] & 20.6\% & 0.1 & [-0.02, 0.21] & 0.252 \\
\midrule
Internal validity (0--5) & o3 & 5.00 [5.00;5.00] & 81.4\% & 5.00 [5.00;5.00] & 78.1\% & 0.04 & [-0.04, 0.13] & 0.849 \\
 & DeepSeek-R1 & 5.00 [4.00;5.00] & 59.2\% & 5.00 [4.00;5.00] & 51.1\% & 0.15 & [0.03, 0.26] & 0.03 \\
 & GPT-4 & 5.00 [4.00;5.00] & 57.8\% & 4.50 [4.00;5.00] & 50.0\% & 0.09 & [-0.02, 0.20] & 0.263 \\
 & Llama-405B & 5.00 [4.00;5.00] & 55.0\% & 4.00 [4.00;5.00] & 42.5\% & 0.26 & [0.14, 0.39] & $<$0.001 \\
 & BioMistral & 2.00 [1.00;3.00] & 4.2\% & 2.00 [1.00;2.00] & 1.1\% & 0.19 & [0.03, 0.36] & 0.054 \\
\midrule
External validity (0--3) & o3 & 3.00 [3.00;3.00] & 96.7\% & 3.00 [3.00;3.00] & 92.5\% & 0.06 & [0.01, 0.10] & 0.0285 \\
 & DeepSeek-R1 & 3.00 [3.00;3.00] & 85.6\% & 3.00 [3.00;3.00] & 86.1\% & 0.003 & [-0.06, 0.07] & 1 \\
 & GPT-4 & 3.00 [3.00;3.00] & 83.3\% & 3.00 [3.00;3.00] & 82.5\% & 0.02 & [-0.05, 0.09] & 1 \\
 & Llama-405B & 3.00 [3.00;3.00] & 83.6\% & 3.00 [2.00;3.00] & 71.9\% & 0.17 & [0.09, 0.25] & $<$0.001 \\
 & BioMistral & 1.00 [1.00;3.00] & 34.4\% & 1.00 [1.00;3.00] & 29.4\% & 0.11 & [-0.02, 0.23] & 0.245 \\
\midrule
Differential diagnosis (0--1) & o3 & 1.00 [1.00;1.00] & 95.0\% & 1.00 [1.00;1.00] & 95.3\% & 0.003 & [-0.03, 0.03] & 1 \\
 & DeepSeek-R1 & 1.00 [0.00;1.00] & 64.7\% & 1.00 [0.00;1.00] & 56.7\% & 0.08 & [0.02, 0.13] & 0.0208 \\
 & GPT-4 & 1.00 [1.00;1.00] & 78.1\% & 1.00 [0.00;1.00] & 65.0\% & 0.13 & [0.07, 0.19] & $<$0.001 \\
 & Llama-405B & 1.00 [1.00;1.00] & 81.7\% & 1.00 [0.00;1.00] & 67.5\% & 0.14 & [0.09, 0.20] & $<$0.001 \\
 & BioMistral & 0.00 [0.00;0.00] & 13.1\% & 0.00 [0.00;0.00] & 10.3\% & 0.03 & [-0.01, 0.07] & 0.492 \\
\midrule
Logical structure (0--4) & o3 & 4.00 [4.00;4.00] & 75.3\% & 4.00 [4.00;4.00] & 79.7\% & -0.08 & [-0.18, 0.02] & 1 \\
 & DeepSeek-R1 & 4.00 [3.00;4.00] & 65.0\% & 4.00 [2.00;4.00] & 60.8\% & 0.1 & [-0.03, 0.23] & 0.339 \\
 & GPT-4 & 4.00 [3.00;4.00] & 71.4\% & 4.00 [3.00;4.00] & 61.9\% & 0.16 & [0.06, 0.26] & 0.00518 \\
 & Llama-405B & 4.00 [3.00;4.00] & 60.6\% & 4.00 [3.00;4.00] & 52.5\% & 0.17 & [0.06, 0.28] & 0.00536 \\
 & BioMistral & 1.00 [1.00;2.00] & 7.2\% & 1.00 [1.00;1.00] & 1.9\% & 0.21 & [0.10, 0.33] & 0.00109 \\
\midrule
Expression (0--2) & o3 & 2.00 [2.00;2.00] & 99.2\% & 2.00 [2.00;2.00] & 93.1\% & 0.06 & [0.04, 0.09] & $<$0.001 \\
 & DeepSeek-R1 & 2.00 [2.00;2.00] & 99.2\% & 2.00 [2.00;2.00] & 97.2\% & 0.02 & [0.00, 0.04] & 0.115 \\
 & GPT-4 & 2.00 [2.00;2.00] & 97.5\% & 2.00 [2.00;2.00] & 94.4\% & 0.03 & [0.00, 0.06] & 0.0845 \\
 & Llama-405B & 2.00 [2.00;2.00] & 99.4\% & 2.00 [2.00;2.00] & 93.6\% & 0.06 & [0.03, 0.08] & $<$0.001 \\
 & BioMistral & 2.00 [2.00;2.00] & 76.1\% & 2.00 [0.00;2.00] & 65.6\% & 0.14 & [0.04, 0.24] & 0.0219 \\
\bottomrule
\end{tabular}
}
\vspace{4pt}
\caption{\textbf{Performance comparison between English and French on evaluation criteria for each model.} Scores are reported as median [interquartile range] and percentage of observations achieving the maximum score (\% max). All scores were treated as continuous in linear mixed models. P-values are one-sided (EN > FR) and Bonferroni-adjusted ($k = 5$).}
\label{tab:table2}
\end{table}

\subsection{Language comparison across clinical contexts}

Exploratory analyses examined whether the language gap varied across clinical contexts: medical specialties, diagnostic reasoning types, and diagnosis types (Supplementary Tables~\ref{tab:table3}, ~\ref{tab:table4}, and~\ref{tab:table5}). Performance across 16 medical specialties was examined descriptively due to the small number of vignettes per specialty (range: 6–32). The language gap was not uniform: for models most affected by language (Llama-405B, GPT-4-Turbo), the largest differences were observed in specialties such as endocrinology, emergency and critical care, and internal medicine, whereas performance remained comparable across languages in cardiovascular medicine and psychiatry. o3 showed stable performance across specialties in both languages.

The language gap also varied by diagnostic reasoning type (Supplementary Table~\ref{tab:table4}). It was most pronounced for hypothetico-deductive (Llama-405B: 1.41, \textit{p}~<~0.001; BioMistral-7B: 1.61, \textit{p}~=~0.002) and algorithmic reasoning (Llama-405B: 1.61, \textit{p}~=~0.003; o3: 0.89, \textit{p}~=~0.017), which require more complex, multi-step inference. Notably, algorithmic reasoning was the only reasoning type in which o3 showed a difference favoring English. Case recognition and forward chaining showed differences primarily for Llama-405B (case recognition: 0.90, \textit{p}~<~0.001; forward chaining: 0.62, \textit{p}~=~0.033), GPT-4-Turbo (0.54, \textit{p}~=~0.015; 0.57, \textit{p}~=~0.045), and DeepSeek-R1 (forward chaining: 0.66, \textit{p}~=~0.037). In contrast, no language effect was observed for probabilistic reasoning for any model.

Regarding diagnosis type, the language gap was most consistent for etiological diagnoses, where three of the five models showed better performance in English (Llama-405B: 0.87, \textit{p}~<~0.001; GPT-4-Turbo: 0.55, \textit{p}~=~0.001; BioMistral-7B: 0.91, \textit{p}~<~0.01). Syndromic diagnoses showed a similar pattern for Llama-405B (1.02, \textit{p}~<~0.001) and GPT-4-Turbo (0.51, \textit{p}~=~0.044), while no differences were observed for ancillary test-based diagnoses, though the small sample size limits interpretation (Supplementary Table~\ref{tab:table5}).

\section{Discussion}

This study shows that prompting language affects both the diagnostic reasoning quality and diagnosis accuracy of large language models. Four of the five models evaluated performed better when prompted in English than in French, with mean differences ranging from 0.37 to 0.91 on the 18-point scale. The language gap was observed across multiple dimensions of clinical reasoning, including differential diagnosis, logical structure, and internal validity, and was consistent across medical specialties, etiological and syndromic diagnosis types. Notably, o3 was the only model to show no overall language effect, suggesting that advances in reasoning capabilities may partially mitigate language-related disparities.

These findings extend prior work documenting multilingual performance gaps in medical LLMs. Previous studies using multiple-choice medical examinations have shown that LLM accuracy declines in non-English settings~\cite{ALONSO2024102938,qiu2024towards,strasser2026performance,yang2026toward}. Our study goes beyond exam-based evaluations by assessing open-ended diagnostic reasoning on clinical vignettes, which is a task closer to real clinical practice, and confirms that the language gap persists in this more critical setting. Moreover, by evaluating both the final diagnosis and the reasoning process leading to it, we show that the gap is not limited to factual recall but extends to the quality and coherence of clinical reasoning itself. This is consistent with recent work suggesting that LLMs trained predominantly on English-language corpora develop stronger reasoning patterns in English, which do not fully transfer to other languages even when the models are capable of generating fluent text in those languages~\cite{etxaniz2024multilingual,wendler2024llamas}.

The finding that o3 showed no language gap on the overall score, while all other models did, warrants attention. Recent work on reasoning language models has shown that test-time compute scaling and extended chain-of-thought generation can improve multilingual reasoning performance, even when reasoning training data is predominantly English~\cite{yong2025crosslingual}. DeepSeek-R1, which also incorporates reinforcement learning-based reasoning, showed the second smallest language gap among models with comparable overall performance. The interaction between model capability and language sensitivity deserves further investigation and attention as training paradigms for reasoning continue to evolve.

From a clinical perspective, these results have implications for the deployment of LLMs in healthcare settings beyond English-speaking practitioners, and by extension non-native English-speaking patients.  If models produce less accurate diagnoses and lower-quality reasoning when prompted in a patient’s native language (other than English), this could exacerbate existing health disparities. Beyond formal diagnostic applications, LLMs are increasingly used by patients directly for health information, making multilingual reliability not merely an academic and hospital concern but a pressing patient safety issue. Given that French is a relatively well-resourced language with substantial representation in training corpora and a long tradition of clinical research, the disparities observed here are likely to be amplified for lower-resource languages~\cite{myung2024blend}. These findings reinforce the need for improvement of large language models in terms of linguistico-cultural alignment, to seize the opportunity to reduce disparities in access to reliable medical information and clinical decision support.

Our evaluation framework, based on six clinical criteria assessed independently by two physicians, offers a replicable methodology for evaluating diagnostic reasoning beyond simple accuracy metrics or Likert scales. While automated evaluation methods are increasingly used for scalability, they struggle to capture the nuanced aspects of clinical reasoning that are central to safe clinical decision-making~\cite{croxford2024development,abacha2023investigation,zhou2025automating}. The moderate inter-rater agreement observed in our study (mean ICC~=~0.48 for the overall score), with higher agreement for lower-performing models and lower agreement for reasoning-specific criteria, reflects the inherent difficulty of evaluating clinical reasoning and is consistent with agreement levels reported in similar medical evaluation studies~\cite{schaekermann2019understanding,schaekermann2018expert}.

To support future research, we openly release the full dataset of 180 bilingual clinical vignettes with physicians' diagnostic reasoning pathways, expected diagnosis, and evaluation scores of the evaluated models (CC-by-NC-4.0 license) . This resource enables benchmarking of LLMs across languages on a clinically grounded task, beyond the multiple-choice format that dominate current medical LLM evaluation. Future work should extend these evaluations to additional languages and clinical contexts.

This study has several limitations. First, our evaluation was limited in scope, as it included only two languages, a small number of models, and standardized clinical vignettes, which may not fully capture the complexity of real-time clinical encounters. The generalizability of the findings to other languages, and particularly low-resource languages, remains to be established. Second, each vignette was queried only once per model and language, without prompt variation, preventing assessment of output variability. Multiple studies have shown the non-determinism of LLMs and the impact of prompt design on the output quality~\cite{mizrahi2024state,song2025good}. Third, inter-rater agreement was moderate for the overall score and low for some sub-scores, reflecting the compound and diverse nature of the reasoning paths in the evaluation. Both physicians were general practitioners of the same age, similar medical training and interest for LLMs, which ensured consistency in assessment standards but may have limited specialist perspectives on domain-specific reasoning patterns. Fourth, the evaluation framework itself had limitations. The binary differential diagnosis criterion lacked the granularity to distinguish between different reasoning approaches. Some models simply eliminated all differential diagnoses by emphasizing negative findings (for example, absence of fever = no meningitis) without structured prioritization. Other models outright listed all possible diagnoses and ruled them out one by one. Default scoring rules, awarding full marks when no differential diagnosis was expected, may have also inflated scores for models generating minimal outputs, such as BioMistral-7B. Similarly, while overt hallucinations were penalized under the expression criterion, more subtle forms of confabulation were observed even in top-performing models and were not systematically captured. A dedicated hallucination criterion and a more granular diagnosis accuracy scale could improve sensitivity to language-related differences in future evaluations. Finally, ceiling effects for high-performing models, particularly o3, limited the ability to detect subtle performance disparities. 

In conclusion, this study provides evidence that prompting language remains a critical determinant of LLM performance in clinical diagnostic reasoning, with consistent advantages for English across most models and evaluation criteria. As LLMs are increasingly considered for clinical applications worldwide, ensuring equitable performance across languages should be a priority for both model development and regulatory evaluation.

\section{Data availability}

The complete dataset of 180 bilingual clinical vignettes (English and French), including physicians' diagnostic reasoning pathways, expected diagnoses, and evaluation scores for all models, is publicly available on Hugging Face (\href{https://huggingface.co/datasets/ANR-MALADES/DiagTrace}{https://huggingface.co/datasets/ANR-MALADES/DiagTrace}) and Zenodo (\href{https://doi.org/10.5281/zenodo.20266783}{https://doi.org/10.5281/zenodo.20266783}).

\section{Code availability}

The code used for statistical analysis, figure generation, and inter-rater agreement computation is publicly available on GitHub (\href{https://github.com/abazoge/DiagTrace}{https://github.com/abazoge/DiagTrace}).

\section*{Acknowledgments}

This work was financially supported by ANR MALADES (ANR-23-IAS1-0005). This work was performed using HPC resources from GENCI-IDRIS (Grant 2024-AD011013715R2).

\section*{Authors Contribution}

A.B., J.C., S.D.S-A. and P-A.G. conceptualized the project. J.C., S.D.S-A., and A.B. performed data acquisition. J.C. and S.D.S-A. performed clinical evaluation and analyses. A.B., J.C., S.D.S-A. drafted the paper. A.B. and P-A.G. supervised the study. All authors reviewed and approved the paper.

\section*{Competing Interests Disclosure}

PA Gourraud is the founder of Methodomics (2008) and its spin-off Big data Santé (2018- “Octopize” brand). He consults for major pharmaceutical companies, and start-ups, all of which are handled through academic pipelines (AstraZeneca, Amgen, Biogen, Boston Scientific, Cemka, Cepton, Cook, Docaposte/Heva, Edimark, Ellipses, Elsevier, Grunenthal, Hemovis, Janssen, IAGE, Lek, Methodomics, Merck, Mérieux, Novartis, Octopize, Sanofi-Genzyme, Lifen, TuneInsight, Aspire UAE). PA Gourraud is a volunteer board member at AXA not-for-profit mutual insurance company (2021). He has no prescription activity with either drugs or devices.  He receives no wages from these activities.
All other authors declare no competing interests.

\newpage

\bibliographystyle{unsrt}
\bibliography{bibtex}


\newpage

\appendix

\section{Supplementary Figures}

\begin{figure}[htbp]
\centering
\includegraphics[width=\linewidth]{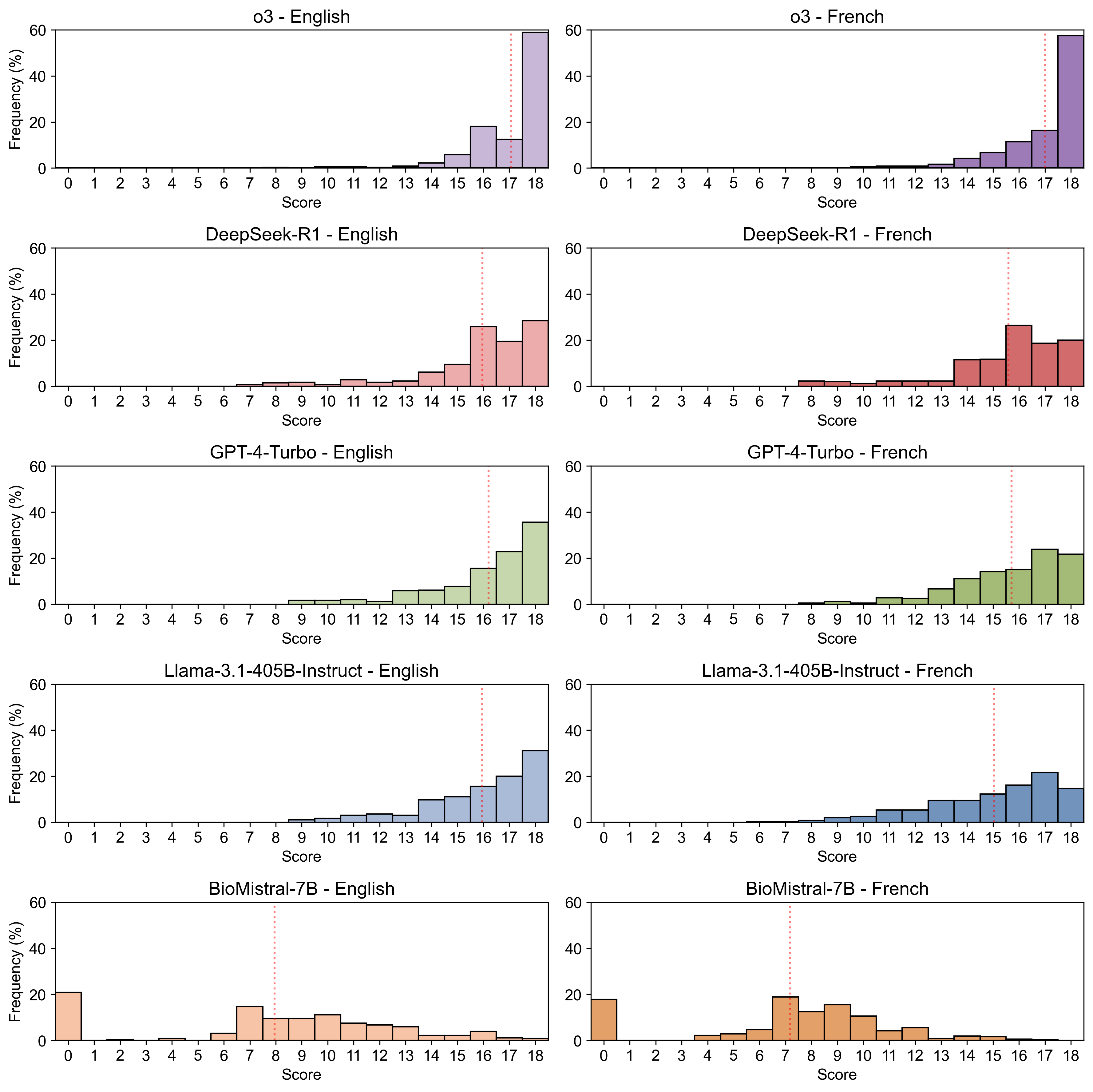}

\vspace{2pt}
\caption{\textbf{Summarized model performances.} Histograms showing the performance of o3, DeepSeek-R1, Llama-3.1-405B-Instruct, GPT-4-Turbo and BioMistral-7B considering prompting of the vignette in French and English. Models are evaluated on a scale of 18 points by two physicians. The red line indicates the mean performance of each model.}
\label{fig:4}
\end{figure}

\newpage

\section{Supplementary Tables}

\begin{table}[h!]
\centering
\resizebox{\textwidth}{!}{%
\begin{tabular}{llccccc}
\toprule
\textbf{Medical specialty} & \textbf{Model} & \textbf{EN median [IQR]} & \textbf{\% max EN} & \textbf{FR median [IQR]} & \textbf{\% max FR} \\
\midrule
General practice (n = 64) & o3 & 18.00 [17.00;18.00] & 64.1\% & 18.00 [17.00;18.00] & 57.8\% \\
 & DeepSeek-R1 & 17.00 [15.75;18.00] & 28.1\% & 17.00 [15.75;18.00] & 28.1\% \\
 & GPT-4-Turbo & 17.00 [16.00;18.00] & 42.2\% & 16.50 [15.00;18.00] & 26.6\% \\
 & Llama-405B & 16.00 [15.00;18.00] & 31.2\% & 15.00 [13.00;17.00] & 15.6\% \\
 & BioMistral-7B & 9.00 [7.00;11.25] & 0.0\% & 8.00 [7.00;10.00] & 0.0\% \\
\midrule
Infectious and tropical diseases & o3 & 18.00 [16.25;18.00] & 55.9\% & 18.00 [16.25;18.00] & 67.6\% \\
(n = 34) & DeepSeek-R1 & 16.00 [15.00;18.00] & 29.4\% & 16.00 [15.25;17.00] & 14.7\% \\
 & GPT-4-Turbo & 17.00 [15.00;18.00] & 29.4\% & 15.50 [14.00;17.00] & 14.7\% \\
 & Llama-405B & 16.00 [15.00;18.00] & 35.3\% & 16.00 [15.00;17.00] & 14.7\% \\
 & BioMistral-7B & 9.00 [7.00;11.00] & 0.0\% & 7.00 [7.00;9.00] & 0.0\% \\
\midrule
Hepatology-gastroenterology & o3 & 18.00 [17.00;18.00] & 66.7\% & 18.00 [17.00;18.00] & 56.7\% \\
(n = 30) & DeepSeek-R1 & 16.00 [15.00;17.00] & 16.7\% & 16.00 [15.00;17.00] & 6.7\% \\
 & GPT-4-Turbo & 16.00 [15.25;18.00] & 30.0\% & 17.00 [15.00;18.00] & 30.0\% \\
 & Llama-405B & 17.00 [15.00;18.00] & 46.7\% & 16.50 [14.25;17.00] & 20.0\% \\
 & BioMistral-7B & 9.50 [7.00;11.00] & 0.0\% & 8.00 [7.00;9.75] & 0.0\% \\
\midrule
Cardiovascular medicine & o3 & 18.00 [16.00;18.00] & 54.2\% & 18.00 [16.75;18.00] & 58.3\% \\
(n = 24) & DeepSeek-R1 & 16.00 [16.00;18.00] & 33.3\% & 17.00 [15.75;18.00] & 37.5\% \\
 & GPT-4-Turbo & 17.00 [15.00;18.00] & 37.5\% & 17.00 [15.00;18.00] & 33.3\% \\
 & Llama-405B & 17.00 [14.00;17.00] & 20.8\% & 15.00 [14.00;17.00] & 20.8\% \\
 & BioMistral-7B & 9.50 [7.50;11.00] & 0.0\% & 8.00 [6.75;9.25] & 0.0\% \\
\midrule
Neurology (n = 22) & o3 & 16.50 [16.00;18.00] & 40.9\% & 18.00 [16.00;18.00] & 54.5\% \\
 & DeepSeek-R1 & 16.00 [15.00;17.00] & 18.2\% & 16.00 [15.00;17.00] & 13.6\% \\
 & GPT-4-Turbo & 17.00 [15.00;18.00] & 36.4\% & 16.00 [15.25;17.00] & 22.7\% \\
 & Llama-405B & 17.00 [12.75;18.00] & 31.8\% & 15.50 [14.00;17.00] & 0.0\% \\
 & BioMistral-7B & 8.50 [7.00;10.75] & 0.0\% & 9.00 [7.00;10.00] & 0.0\% \\
\midrule
Pulmunology (n = 20) & o3 & 17.50 [15.75;18.00] & 50.0\% & 17.50 [15.00;18.00] & 50.0\% \\
 & DeepSeek-R1 & 16.00 [15.00;17.00] & 20.0\% & 16.00 [15.00;17.00] & 20.0\% \\
 & GPT-4-Turbo & 18.00 [16.75;18.00] & 60.0\% & 17.00 [16.00;17.00] & 20.0\% \\
 & Llama-405B & 17.00 [15.75;18.00] & 45.0\% & 17.00 [16.00;18.00] & 45.0\% \\
 & BioMistral-7B & 9.50 [0.00;13.00] & 5.0\% & 10.00 [6.75;12.50] & 0.0\% \\
\midrule
Emergency and critical care & o3 & 18.00 [16.00;18.00] & 55.0\% & 18.00 [16.00;18.00] & 65.0\% \\
(n = 20) & DeepSeek-R1 & 17.00 [16.00;18.00] & 35.0\% & 16.00 [15.00;17.25] & 25.0\% \\
 & GPT-4-Turbo & 16.50 [16.00;17.25] & 25.0\% & 14.50 [14.00;16.25] & 20.0\% \\
 & Llama-405B & 16.50 [15.00;18.00] & 30.0\% & 15.00 [11.00;17.00] & 10.0\% \\
 & BioMistral-7B & 3.00 [0.00;9.00] & 0.0\% & 8.00 [6.75;9.00] & 0.0\% \\
\midrule
Endocrinology and metabolism & o3 & 17.50 [17.00;18.00] & 50.0\% & 17.00 [16.25;18.00] & 38.9\% \\
(n = 18) & DeepSeek-R1 & 16.00 [16.00;17.75] & 27.8\% & 16.00 [14.00;16.00] & 16.7\% \\
 & GPT-4-Turbo & 16.50 [15.00;17.00] & 22.2\% & 16.50 [14.00;17.00] & 11.1\% \\
 & Llama-405B & 17.00 [16.00;18.00] & 33.3\% & 14.00 [11.50;16.75] & 5.6\% \\
 & BioMistral-7B & 7.00 [6.25;10.00] & 0.0\% & 7.00 [5.00;9.00] & 0.0\% \\
\midrule
Gynecology (n = 18) & o3 & 18.00 [16.00;18.00] & 61.1\% & 18.00 [17.00;18.00] & 66.7\% \\
 & DeepSeek-R1 & 18.00 [16.25;18.00] & 55.6\% & 16.50 [15.25;18.00] & 33.3\% \\
 & GPT-4-Turbo & 17.00 [16.00;17.75] & 27.8\% & 16.50 [15.00;17.00] & 22.2\% \\
 & Llama-405B & 16.00 [14.50;17.00] & 11.1\% & 16.50 [14.25;17.00] & 11.1\% \\
 & BioMistral-7B & 9.50 [8.00;12.75] & 0.0\% & 9.00 [5.50;10.00] & 0.0\% \\
\midrule
Oncology-hematology (n = 18) & o3 & 17.50 [16.00;18.00] & 50.0\% & 17.00 [16.00;18.00] & 38.9\% \\
 & DeepSeek-R1 & 17.00 [16.00;18.00] & 33.3\% & 15.50 [13.25;16.00] & 11.1\% \\
 & GPT-4-Turbo & 17.00 [17.00;18.00] & 38.9\% & 16.50 [15.25;17.75] & 27.8\% \\
 & Llama-405B & 17.00 [13.25;18.00] & 38.9\% & 16.00 [11.25;17.00] & 11.1\% \\
 & BioMistral-7B & 8.00 [7.25;10.00] & 0.0\% & 7.00 [4.25;8.00] & 0.0\% \\
\midrule
Rheumatology (n = 18) & o3 & 18.00 [17.25;18.00] & 72.2\% & 18.00 [18.00;18.00] & 77.8\% \\
 & DeepSeek-R1 & 17.00 [16.00;18.00] & 33.3\% & 14.50 [13.25;16.00] & 11.1\% \\
 & GPT-4-Turbo & 17.00 [16.00;17.00] & 22.2\% & 15.50 [14.25;17.00] & 0.0\% \\
 & Llama-405B & 16.00 [15.00;17.00] & 22.2\% & 14.50 [13.00;16.75] & 11.1\% \\
 & BioMistral-7B & 7.00 [7.00;9.00] & 0.0\% & 5.00 [0.00;8.75] & 0.0\% \\
\midrule
Urology-Nephrology (n = 18) & o3 & 18.00 [18.00;18.00] & 77.8\% & 17.00 [15.25;18.00] & 38.9\% \\
 & DeepSeek-R1 & 16.00 [15.00;17.75] & 27.8\% & 14.50 [14.00;16.00] & 16.7\% \\
 & GPT-4-Turbo & 17.00 [15.00;18.00] & 38.9\% & 15.00 [11.25;16.75] & 16.7\% \\
 & Llama-405B & 16.00 [14.00;17.00] & 16.7\% & 14.00 [12.25;16.75] & 11.1\% \\
 & BioMistral-7B & 9.50 [0.00;11.75] & 0.0\% & 6.50 [0.00;8.75] & 0.0\% \\
\bottomrule
\end{tabular}
}
\vspace{4pt}
\parbox{\textwidth}{\footnotesize \textit{Continued on next page.}}
\end{table}

\newpage

\begin{table}[t!]
\centering
\resizebox{\textwidth}{!}{%
\begin{tabular}{llccccc}
\toprule
\textbf{Medical specialty} & \textbf{Model} & \textbf{EN median [IQR]} & \textbf{\% max EN} & \textbf{FR median [IQR]} & \textbf{\% max FR} \\
\midrule
Psychiatry (n = 16) & o3 & 17.00 [16.00;18.00] & 43.8\% & 17.50 [17.00;18.00] & 50.0\% \\
 & DeepSeek-R1 & 16.00 [16.00;18.00] & 31.2\% & 16.00 [15.00;16.25] & 18.8\% \\
 & GPT-4-Turbo & 16.00 [14.00;17.25] & 25.0\% & 15.00 [14.00;16.25] & 6.2\% \\
 & Llama-405B & 16.00 [15.00;16.00] & 12.5\% & 16.00 [15.00;17.00] & 6.2\% \\
 & BioMistral-7B & 4.50 [0.00;11.00] & 0.0\% & 9.50 [8.00;10.25] & 0.0\% \\
\midrule
Pediatrics (n = 14) & o3 & 18.00 [16.25;18.00] & 64.3\% & 18.00 [18.00;18.00] & 78.6\% \\
 & DeepSeek-R1 & 16.00 [16.00;17.00] & 21.4\% & 15.50 [15.00;16.00] & 14.3\% \\
 & GPT-4-Turbo & 17.00 [14.25;17.75] & 28.6\% & 17.00 [14.00;18.00] & 42.9\% \\
 & Llama-405B & 16.00 [15.00;18.00] & 42.9\% & 15.00 [13.00;17.00] & 14.3\% \\
 & BioMistral-7B & 10.00 [1.75;11.50] & 14.3\% & 0.00 [0.00;5.25] & 0.0\% \\
\midrule
Head and neck medicine & o3 & 18.00 [18.00;18.00] & 78.6\% & 18.00 [17.00;18.00] & 64.3\% \\
(n = 14) & DeepSeek-R1 & 15.00 [11.75;16.75] & 14.3\% & 16.00 [11.25;16.75] & 14.3\% \\
 & GPT-4-Turbo & 17.50 [14.50;18.00] & 50.0\% & 14.50 [11.25;17.00] & 7.1\% \\
 & Llama-405B & 17.00 [14.50;18.00] & 35.7\% & 15.50 [13.50;16.75] & 21.4\% \\
 & BioMistral-7B & 8.00 [7.00;12.50] & 0.0\% & 7.00 [6.25;8.75] & 0.0\% \\
\midrule
Internal medicine (n = 12) & o3 & 17.50 [16.00;18.00] & 50.0\% & 17.50 [16.50;18.00] & 50.0\% \\
 & DeepSeek-R1 & 16.50 [15.00;18.00] & 33.3\% & 16.50 [16.00;17.25] & 25.0\% \\
 & GPT-4-Turbo & 17.50 [16.50;18.00] & 50.0\% & 16.50 [14.50;18.00] & 33.3\% \\
 & Llama-405B & 17.00 [15.75;18.00] & 33.3\% & 14.50 [12.75;17.00] & 8.3\% \\
 & BioMistral-7B & 3.00 [0.00;8.00] & 0.0\% & 7.00 [6.50;9.00] & 0.0\% \\
\bottomrule
\end{tabular}
}
\vspace{4pt}
\caption{\textbf{Descriptive performance by medical specialty.} \textit{n} refers to the number of paired assessments (each vignette assessed by two independent physicians). Scores are reported as median [interquartile range] and percentage achieving the maximum score on the overall scale (0--18). No inferential statistics were performed due to the small number of vignettes per specialty (ranging from 6 to 32).}
\label{tab:table3}
\end{table}

\begin{table}[htbp]
\centering
\resizebox{\textwidth}{!}{%
\begin{tabular}{llcccccccc}
\toprule
\textbf{Diagnostic reasoning} & \textbf{Model} & \textbf{EN median [IQR]} & \textbf{\% max} & \textbf{FR median [IQR]} & \textbf{\% max} & \textbf{Mean diff} & \textbf{95\% CI} & \textbf{Adj. \textit{p}} \\
 &  &  & \textbf{EN} &  & \textbf{FR} & \textbf{(LMM)} &  &  \\
\midrule
Case recognition & o3 & 18.00 [16.00;18.00] & 54.4\% & 18.00 [16.00;18.00] & 57.9\% & 0.0 & [-0.36, 0.36] & 1 \\
(n = 114) & DeepSeek-R1 & 17.00 [16.00;18.00] & 34.2\% & 16.00 [15.00;17.00] & 21.9\% & 0.2 & [-0.26, 0.66] & 0.981 \\
 & GPT-4-Turbo & 17.00 [15.00;18.00] & 37.7\% & 16.00 [14.00;17.00] & 21.9\% & 0.54 & [0.16, 0.93] & 0.0146 \\
 & Llama-405B & 16.00 [15.00;18.00] & 36.0\% & 16.00 [14.00;17.00] & 18.4\% & 0.9 & [0.48, 1.33] & $<$0.001 \\
 & BioMistral-7B & 9.00 [7.00;12.00] & 1.8\% & 8.00 [7.00;10.00] & 0.0\% & 0.47 & [-0.49, 1.44] & 0.844 \\
\addlinespace[3pt]
Forward chaining & o3 & 18.00 [16.00;18.00] & 54.5\% & 18.00 [17.00;18.00] & 56.4\% & -0.17 & [-0.51, 0.16] & 1 \\
(n = 110) & DeepSeek-R1 & 16.00 [16.00;17.00] & 20.9\% & 16.00 [14.00;17.00] & 17.3\% & 0.66 & [0.13, 1.19] & 0.0373 \\
 & GPT-4-Turbo & 17.00 [16.00;18.00] & 30.9\% & 16.50 [14.00;17.00] & 16.4\% & 0.57 & [0.10, 1.04] & 0.0449 \\
 & Llama-405B & 16.00 [14.00;17.00] & 22.7\% & 16.00 [13.00;17.00] & 10.9\% & 0.62 & [0.13, 1.10] & 0.0329 \\
 & BioMistral-7B & 8.00 [0.00;10.00] & 0.0\% & 7.00 [1.00;9.00] & 0.0\% & 0.63 & [-0.35, 1.60] & 0.522 \\
\addlinespace[3pt]
Hypothetico-deductive & o3 & 18.00 [16.00;18.00] & 60.8\% & 18.00 [16.25;18.00] & 60.8\% & 0.14 & [-0.31, 0.58] & 1 \\
(n = 74) & DeepSeek-R1 & 16.00 [15.00;18.00] & 29.7\% & 16.00 [15.00;17.00] & 20.3\% & 0.24 & [-0.30, 0.79] & 0.961 \\
 & GPT-4-Turbo & 17.00 [15.25;18.00] & 35.1\% & 16.50 [15.00;18.00] & 27.0\% & 0.31 & [-0.18, 0.81] & 0.553 \\
 & Llama-405B & 17.00 [15.00;18.00] & 36.5\% & 15.00 [13.25;17.00] & 12.2\% & 1.41 & [0.84, 1.97] & $<$0.001 \\
 & BioMistral-7B & 10.00 [7.00;12.00] & 1.4\% & 8.00 [7.00;10.00] & 0.0\% & 1.61 & [0.70, 2.51] & 0.00175 \\
\addlinespace[3pt]
Probabilistic & o3 & 18.00 [17.00;18.00] & 70.6\% & 18.00 [17.00;18.00] & 64.7\% & 0.32 & [-0.19, 0.84] & 0.561 \\
(n = 34) & DeepSeek-R1 & 16.00 [15.00;18.00] & 29.4\% & 17.00 [16.00;17.00] & 23.5\% & 0.15 & [-0.82, 1.11] & 1 \\
 & GPT-4-Turbo & 17.00 [16.00;18.00] & 38.2\% & 16.00 [14.00;17.75] & 26.5\% & 0.85 & [-0.05, 1.76] & 0.178 \\
 & Llama-405B & 17.00 [15.00;18.00] & 35.3\% & 16.00 [14.25;17.75] & 26.5\% & 0.26 & [-0.64, 1.17] & 1 \\
 & BioMistral-7B & 9.00 [7.00;11.00] & 0.0\% & 9.00 [7.00;10.00] & 0.0\% & -0.12 & [-1.46, 1.22] & 1 \\
\addlinespace[3pt]
Algorithmic & o3 & 18.00 [17.75;18.00] & 75.0\% & 17.00 [15.75;18.00] & 42.9\% & 0.89 & [0.28, 1.51] & 0.0172 \\
(n = 28) & DeepSeek-R1 & 16.00 [15.00;18.00] & 28.6\% & 16.00 [14.00;17.00] & 17.9\% & 0.46 & [-0.48, 1.41] & 0.852 \\
 & GPT-4-Turbo & 17.00 [14.75;18.00] & 42.9\% & 16.00 [15.00;17.00] & 21.4\% & -0.0 & [-0.95, 0.95] & 1 \\
 & Llama-405B & 16.00 [15.00;17.25] & 25.0\% & 14.00 [13.00;17.00] & 7.1\% & 1.61 & [0.71, 2.50] & 0.00262 \\
 & BioMistral-7B & 8.50 [0.00;10.00] & 0.0\% & 7.00 [0.00;9.00] & 0.0\% & 1.5 & [-0.27, 3.27] & 0.262 \\
\bottomrule
\end{tabular}
}
\vspace{4pt}
\caption{\textbf{Performance comparison between English and French by diagnostic reasoning type.} \textit{n} refers to the number of paired assessments (each vignette assessed by two independent physicians). Scores are reported as median [interquartile range] and percentage achieving the maximum score on the overall scale (0--18). Differences were assessed using linear mixed models. P-values are one-sided (EN > FR) and Bonferroni-adjusted ($k = 5$).}
\label{tab:table4}
\end{table}

\begin{table}[t!]
\centering
\resizebox{\textwidth}{!}{%
\begin{tabular}{llcccccccc}
\toprule
\textbf{Diagnosis} & \textbf{Model} & \textbf{EN median [IQR]} & \textbf{\% max} & \textbf{FR median [IQR]} & \textbf{\% max} & \textbf{Mean diff} & \textbf{95\% CI} & \textbf{Adj. \textit{p}} \\
 &  & & \textbf{EN} &  & \textbf{FR} & \textbf{(LMM)} &  &  \\
\midrule
Etiologic & o3 & 18.00 [17.00;18.00] & 62.4\% & 18.00 [16.00;18.00] & 56.6\% & 0.26 & [0.00, 0.51] & 0.119 \\
(n = 226) & DeepSeek-R1 & 16.00 [15.00;18.00] & 29.2\% & 16.00 [15.00;17.00] & 19.5\% & 0.31 & [-0.03, 0.66] & 0.194 \\
 & GPT-4-Turbo & 17.00 [15.00;18.00] & 38.1\% & 16.00 [14.00;17.00] & 22.6\% & 0.55 & [0.24, 0.86] & 0.0013 \\
 & Llama-405B & 17.00 [15.00;18.00] & 32.3\% & 16.00 [14.00;17.00] & 17.3\% & 0.87 & [0.53, 1.20] & $<$0.001 \\
 & BioMistral-7B & 9.00 [7.00;11.00] & 0.4\% & 8.00 [6.00;9.00] & 0.0\% & 0.91 & [0.30, 1.52] & 0.00881 \\
\addlinespace[3pt]
Syndromic & o3 & 18.00 [16.00;18.00] & 53.7\% & 18.00 [17.00;18.00] & 59.3\% & -0.21 & [-0.51, 0.09] & 1 \\
(n = 108) & DeepSeek-R1 & 16.00 [16.00;18.00] & 29.6\% & 16.00 [15.00;17.00] & 24.1\% & 0.44 & [-0.04, 0.91] & 0.188 \\
 & GPT-4-Turbo & 17.00 [16.00;18.00] & 34.3\% & 16.00 [15.00;17.00] & 19.4\% & 0.51 & [0.09, 0.93] & 0.0441 \\
 & Llama-405B & 17.00 [15.00;18.00] & 32.4\% & 16.00 [13.00;17.00] & 11.1\% & 1.02 & [0.57, 1.47] & $<$0.001 \\
 & BioMistral-7B & 8.00 [6.00;11.00] & 1.9\% & 8.00 [5.75;10.00] & 0.0\% & 0.48 & [-0.51, 1.47] & 0.857 \\
\addlinespace[3pt]
Paraclinical & o3 & 17.50 [16.00;18.00] & 50.0\% & 18.00 [16.25;18.00] & 57.7\% & -0.31 & [-1.10, 0.49] & 1 \\
(n = 26) & DeepSeek-R1 & 16.00 [15.00;17.00] & 15.4\% & 15.50 [13.25;16.75] & 7.7\% & 0.54 & [-0.50, 1.57] & 0.786 \\
 & GPT-4-Turbo & 16.00 [15.00;17.00] & 19.2\% & 16.50 [14.00;17.00] & 23.1\% & -0.08 & [-1.13, 0.97] & 1 \\
 & Llama-405B & 16.00 [13.25;17.00] & 15.4\% & 14.50 [12.00;16.75] & 7.7\% & 0.88 & [0.02, 1.75] & 0.13 \\
 & BioMistral-7B & 7.00 [0.00;11.50] & 0.0\% & 7.00 [0.00;8.00] & 0.0\% & 0.85 & [-0.98, 2.68] & 0.927 \\
\bottomrule
\end{tabular}
}
\vspace{4pt}
\caption{\textbf{Performance comparison between English and French by diagnosis type.} \textit{n} refers to the number of paired assessments (each vignette assessed by two independent physicians). Scores are reported as median [interquartile range] and percentage achieving the maximum score on the overall scale (0--18). Differences were assessed using linear mixed models. P-values are one-sided (EN > FR) and Bonferroni-adjusted ($k = 5$).}
\label{tab:table5}
\end{table}

\begin{table}[htbp]
\centering
\resizebox{\textwidth}{!}{%
\begin{tabular}{llccc}
\toprule
\textbf{Score} & \textbf{Model} & \textbf{Method} & \textbf{EN Value [95\% CI]} & \textbf{FR Value [95\% CI]} \\
\midrule
Overall score (0--18) & o3 & ICC & 0.22 [0.09, 0.35] & 0.15 [0.04, 0.26] \\
& DeepSeek-R1 & ICC & 0.37 [0.22, 0.50] & 0.40 [0.22, 0.54] \\
 & GPT-4-Turbo & ICC & 0.31 [0.17, 0.45] & 0.52 [0.39, 0.63] \\
 & Llama-405B & ICC & 0.61 [0.49, 0.70] & 0.56 [0.44, 0.65] \\
 & BioMistral-7B & ICC & 0.82 [0.76, 0.86] & 0.84 [0.78, 0.88] \\
\midrule
Final diagnosis (0--3) & o3 & Weighted $\kappa$ & 0.38 [0.20, 0.54] & 0.32 [0.12, 0.51] \\
 & DeepSeek-R1 & Weighted $\kappa$ & 0.56 [0.39, 0.70] & 0.62 [0.47, 0.75] \\
 & GPT-4 & Weighted $\kappa$ & 0.26 [0.13, 0.40] & 0.73 [0.62, 0.82] \\
 & Llama-405B & Weighted $\kappa$ & 0.73 [0.62, 0.81] & 0.71 [0.61, 0.80] \\
 & BioMistral & Weighted $\kappa$ & 0.89 [0.84, 0.93] & 0.92 [0.87, 0.95] \\
\midrule
Internal validity (0--5) & o3 & Weighted $\kappa$ & -0.05 [-0.11, 0.03] & -0.04 [-0.17, 0.09] \\
 & DeepSeek-R1 & Weighted $\kappa$ & 0.08 [-0.05, 0.22] & 0.24 [0.03, 0.41] \\
 & GPT-4-Turbo & Weighted $\kappa$ & 0.25 [0.05, 0.44] & 0.25 [0.08, 0.42] \\
 & Llama-405B & Weighted $\kappa$ & 0.29 [0.08, 0.46] & 0.26 [0.11, 0.41] \\
 & BioMistral-7B & Weighted $\kappa$ & 0.62 [0.51, 0.71] & 0.55 [0.43, 0.65] \\
\midrule
External validity (0--3) & o3 & Weighted $\kappa$ & 0.27 [-0.02, 0.60] & -0.06 [-0.09, -0.03] \\
 & DeepSeek-R1 & Weighted $\kappa$ & 0.10 [-0.08, 0.30] & 0.12 [-0.06, 0.31] \\
 & GPT-4-Turbo& Weighted $\kappa$ & 0.06 [-0.08, 0.22] & 0.11 [-0.03, 0.26] \\
 & Llama-405B & Weighted $\kappa$ & 0.29 [0.09, 0.47] & 0.05 [-0.08, 0.18] \\
 & BioMistral-7B & Weighted $\kappa$ & 0.39 [0.31, 0.46] & 0.28 [0.20, 0.35] \\
\midrule
Differential diagnosis (0--1) & o3 & $\kappa$ & -0.02 [-0.04, 0.00] & 0.00 [0.00, 0.00] \\
 & DeepSeek-R1 & $\kappa$ & 0.05 [-0.04, 0.13] & 0.10 [0.03, 0.18] \\
 & GPT-4-Turbo & $\kappa$ & 0.05 [-0.09, 0.20] & 0.32 [0.17, 0.45] \\
 & Llama-405B & $\kappa$ & 0.15 [-0.01, 0.31] & 0.23 [0.08, 0.38] \\
 & BioMistral-7B & $\kappa$ & 0.10 [-0.06, 0.28] & 0.25 [0.04, 0.46] \\
\midrule
Logical structure (0--4) & o3 & Weighted $\kappa$ & 0.11 [0.03, 0.21] & 0.15 [0.03, 0.27] \\
 & DeepSeek-R1 & Weighted $\kappa$ & 0.16 [0.03, 0.30] & 0.09 [-0.02, 0.20] \\
 & GPT-4-Turbo & Weighted $\kappa$ & 0.11 [-0.03, 0.26] & 0.21 [0.09, 0.33] \\
 & Llama-405B & Weighted $\kappa$ & 0.27 [0.13, 0.41] & 0.23 [0.10, 0.36] \\
 & BioMistral-7B & Weighted $\kappa$ & 0.41 [0.32, 0.50] & 0.47 [0.37, 0.56] \\
\midrule
Expression (0--2) & o3 & Weighted $\kappa$ & 0.00 [0.00, 0.00] & 0.13 [-0.03, 0.33] \\
 & DeepSeek-R1 & Weighted $\kappa$ & -0.01 [-0.02, 0.00] & -0.03 [-0.05, -0.01] \\
 & GPT-4-Turbo & Weighted $\kappa$ & -0.01 [-0.03, 0.00] & -0.02 [-0.05, 0.00] \\
 & Llama-405B & Weighted $\kappa$ & 0.00 [0.00, 0.00] & 0.12 [-0.06, 0.34] \\
 & BioMistral-7B & Weighted $\kappa$ & 0.82 [0.71, 0.91] & 0.53 [0.40, 0.65] \\
\bottomrule
\end{tabular}
}
\vspace{4pt}
\caption{\textbf{Inter-rater agreement.} Inter-rater reliability between two physicians evaluators across all models and languages. The overall score (0--18) was assessed using the intraclass correlation coefficient (ICC, two-way random, single measures, absolute agreement). Ordinal sub-scores were assessed using quadratic-weighted Cohen’s kappa, and the binary sub-score (differential diagnosis) using unweighted Cohen’s kappa. 95\% confidence intervals were computed by bootstrap (2,000 resamples).}
\label{tab:table6}
\end{table}

\begin{table}[htbp]
\centering
\resizebox{\textwidth}{!}{%
\begin{tabular}{lccccccccccc}
\toprule
\textbf{Model} & \textbf{EN median [IQR]} & \textbf{FR median [IQR]} & \textbf{V} & \textbf{Hodges-} & \textbf{95\% CI} & \textbf{Rank-} & \textbf{EN > FR} & \textbf{FR > EN} & \textbf{Ties} & \textbf{Adj. \textit{p}} \\
 &  &  &  & \textbf{Lehmann} &  & \textbf{biserial r} &  &  &  & \\
\midrule
o3 & 17.50 [16.50;18.00] & 17.50 [16.50;18.00] & 4211 & 0.0 & [-0.25, Inf] & 0.02 & 65 & 62 & 53 & 1 \\
\addlinespace[3pt]
DeepSeek-R1 & 16.50 [15.50;17.12] & 16.00 [15.00;17.00] & 7743 & 0.5 & [0.25, Inf] & 0.25 & 96 & 57 & 27 & 0.00174 \\
\addlinespace[3pt]
GPT-4-Turbo & 16.50 [15.50;17.50] & 16.00 [15.00;17.00] & 7579 & 0.75 & [0.50, Inf] & 0.4 & 102 & 44 & 34 & $<$0.001 \\
\addlinespace[3pt]
Llama-405B & 16.50 [15.50;17.50] & 15.50 [13.50;16.50] & 9067 & 1.0 & [0.75, Inf] & 0.46 & 112 & 41 & 27 & $<$0.001 \\
\addlinespace[3pt]
BioMistral-7B & 9.00 [6.50;11.00] & 8.00 [6.50;9.50] & 7665 & 1.0 & [0.50, Inf] & 0.26 & 99 & 58 & 23 & 0.0257 \\
\bottomrule
\end{tabular}
}
\vspace{4pt}
\caption{\textbf{Sensitivity analysis (overall score).} Paired Wilcoxon signed-rank tests comparing English and French performance on the overall score (0--18). Scores were aggregated by averaging the two raters’ evaluations per vignette ($n = 180$). P-values are one-sided (EN > FR) and Bonferroni-adjusted ($k = 5$). V = Wilcoxon test statistic; Hodges-Lehmann = pseudo-median of pairwise differences; rank-biserial r = effect size.}
\label{tab:table7}
\end{table}

\begin{table}[htbp]
\centering
\resizebox{\textwidth}{!}{%
\begin{tabular}{llcccccc}
\toprule
\textbf{Criterion (scale)} & \textbf{Model} & \textbf{EN median [IQR]} & \textbf{FR median [IQR]} & \textbf{V} & \textbf{Hodges-} & \textbf{Adj. \textit{p}} \\
 &  &  &  & & \textbf{Lehmann} &  \\
\midrule
Final diagnosis (0--3) & o3 & 3.00 [3.00;3.00] & 3.00 [3.00;3.00] & 460 & -0.0 & 1 \\
 & DeepSeek-R1 & 3.00 [2.50;3.00] & 3.00 [2.50;3.00] & 1122 & 0.0 & 1 \\
 & GPT-4-Turbo & 3.00 [2.50;3.00] & 3.00 [2.50;3.00] & 1260 & 0.25 & 0.526 \\
 & Llama-405B & 3.00 [2.00;3.00] & 3.00 [2.00;3.00] & 1428 & 0.25 & 0.0915 \\
 & BioMistral-7B & 1.00 [1.00;2.50] & 1.00 [1.00;2.00] & 3023 & 0.25 & 0.634 \\
\midrule
Internal validity (0--5) & o3 & 5.00 [4.50;5.00] & 5.00 [4.50;5.00] & 0.0 & 0.87 \\
 & DeepSeek-R1 & 4.50 [4.50;5.00] & 4.50 [4.00;5.00] & 3259 & 0.25 & 0.0247 \\
 & GPT-4-Turbo & 4.50 [4.00;5.00] & 4.50 [4.00;5.00] & 3246 & 0.25 & 0.0819 \\
 & Llama-405B & 4.50 [4.00;5.00] & 4.50 [4.00;4.50] & 4042 & 0.5 & $<$0.001 \\
 & BioMistral-7B & 2.00 [1.50;3.00] & 1.50 [1.50;2.50] & 5616 & 0.25 & 0.282 \\
\midrule
External validity (0--3) & o3 & 3.00 [3.00;3.00] & 3.00 [3.00;3.00] & 373 & 0.5 & 0.0263 \\
 & DeepSeek-R1 & 3.00 [2.50;3.00] & 3.00 [3.00;3.00] & 894 & 0.0 & 1 \\
 & GPT-4-Turbo & 3.00 [2.50;3.00] & 3.00 [2.50;3.00] & 1229 & 0.0 & 1 \\
 & Llama-405B & 3.00 [2.50;3.00] & 3.00 [2.50;3.00] & 2553 & 0.5 & $<$0.001 \\
 & BioMistral-7B & 2.00 [1.00;2.00] & 2.00 [1.00;2.00] & 3320 & 0.25 & 0.609 \\
\midrule
Differential diagnosis (0--1) & o3 & 1.00 [1.00;1.00] & 1.00 [1.00;1.00] & 121 & 0.0 & 1 \\
 & DeepSeek-R1 & 0.50 [0.50;1.00] & 0.50 [0.50;1.00] & 1512 & 0.25 & 0.0108 \\
 & GPT-4-Turbo & 1.00 [0.50;1.00] & 0.50 [0.50;1.00] & 1864 & 0.5 & $<$0.001 \\
 & Llama-405B & 1.00 [0.50;1.00] & 1.00 [0.50;1.00] & 2065 & 0.5 & $<$0.001 \\
 & BioMistral-7B & 0.00 [0.00;0.00] & 0.00 [0.00;0.00] & 597 & 0.0 & 0.471 \\
\midrule
Logical structure (0--4) & o3 & 4.00 [3.00;4.00] & 4.00 [3.50;4.00] & 1657 & -0.0 & 1 \\
 & DeepSeek-R1 & 3.50 [3.00;4.00] & 3.00 [3.00;4.00] & 4162 & 0.0 & 0.27 \\
 & GPT-4-Turbo & 4.00 [3.00;4.00] & 3.50 [3.00;4.00] & 2609 & 0.25 & 0.00286 \\
 & Llama-405B & 3.50 [3.00;4.00] & 3.00 [3.00;4.00] & 3612 & 0.25 & 0.00917 \\
 & BioMistral-7B & 1.00 [1.00;2.00] & 1.00 [1.00;1.50] & 4263 & 0.25 & 0.0222 \\
\midrule
Expression (0--2) & o3 & 2.00 [2.00;2.00] & 2.00 [2.00;2.00] & 231 & 0.5 & $<$0.001 \\
 & DeepSeek-R1 & 2.00 [2.00;2.00] & 2.00 [2.00;2.00] & 54 & 0.5 & 0.0984 \\
 & GPT-4-Turbo & 2.00 [2.00;2.00] & 2.00 [2.00;2.00] & 204 & 0.5 & 0.0571 \\
 & Llama-405B & 2.00 [2.00;2.00] & 2.00 [2.00;2.00] & 254 & 0.5 & $<$0.001 \\
 & BioMistral-7B & 2.00 [1.00;2.00] & 2.00 [1.00;2.00] & 2771 & 0.25 & 0.252 \\
\bottomrule
\end{tabular}
}
\vspace{4pt}
\caption{\textbf{Sensitivity analysis (sub-scores).} Paired Wilcoxon signed-rank tests on aggregated scores ($n = 180$) for each evaluation criterion. P-values are one-sided (EN > FR) and Bonferroni-adjusted ($k = 5$).}
\label{tab:table8}
\end{table}

\begin{table}[t!]
\centering
\resizebox{\textwidth}{!}{%
\begin{tabular}{p{2.5cm} p{5.5cm} p{5.8cm}}
\toprule
\textbf{Criterion (scale)} & \textbf{Description} & \textbf{Scoring} \\
\midrule
Internal validity \newline
(0--5) & Assesses the model's ability to extract and interpret relevant clinical data from the vignette, including translation of semiological descriptions into appropriate medical terminology (e.g., recognizing purpura from "purlish skin lesion that does not fade under pressure") and contextualization of numerical (e.g., 38.5°C as fever). & 
5: Exhaustive and accurate interpretation; elements prioritized. \newline
4: Incomplete but sufficient for diagnosis; some irrelevant elements noted. \newline
3: Incomplete and insufficient for diagnosis. \newline
2: Analysis leads to incorrect diagnosis. \newline
1: Paraphrases the case without analysis. \newline
0: Hallucination or no response.  \\
\midrule
External validity \newline
(0--3) & Assesses the accuracy of medical knowledge introduced beyond the vignette description. Default score of 3 if no external knowledge is invoked. & 
3: Correct and useful for diagnosis. \newline
2: Partially correct or outdated. \newline
1: Incorrect. \newline
0: No response.  \\
\midrule
Differential \newline diagnosis \newline
(0--1) & Assesses the ability to formulate relevant hypothesis and differential diagnoses, ideally prioritized by likelihood or severity. Score of 0 if expected differentials are omitted; default score of 1 if the vignette does not warrant differentials. & 
1: Relevant differential diagnoses suggested and prioritized. \newline
0: No relevant differentials suggested.  \\
\midrule
Logical structure \newline
(0--4) & Evaluates the coherence and organization of reasoning, independently of content accuracy. Attention to logical ordering, absence of contradiction, and exclusion of irrelevant assertions. & 
4: Coherent reasoning oriented toward correct diagnosis. \newline
3: Coherent reasoning oriented toward incorrect diagnosis. \newline
2: Incoherent reasoning with contradictions. \newline
1: No logical connection between elements. \newline
0: No response.  \\
\midrule
Expression \newline
(0--2) & Penalizes errors in expression, meaning or syntax. Also penalizes responses generated in English when French is expected or when the model is not able to generate an answer.  & 
2: No syntactic errors or incorrect phrasing. \newline
1: Language errors not altering meaning. \newline
0: Errors altering meaning or no response.  \\
\midrule
Final diagnosis \newline
(0--3) & Compares the model’s diagnosis with the physician reference diagnosis. & 
3: Perfect diagnosis. \newline
2: Correct but incomplete (e.g., syndromic, non-lateralized, or lacking severity). \newline
1: Incorrect diagnosis. \newline
0: No response.  \\
\bottomrule
\end{tabular}
}
\vspace{4pt}
\caption{\textbf{Detailed evaluation scoring rubric.}}
\label{tab:table9}
\end{table}

\end{document}